\documentclass[sigconf]{acmart}

\usepackage{booktabs} % For formal tables
\usepackage{lipsum}
\usepackage{multirow}
\usepackage{array}
\usepackage{courier}
\usepackage{mathtools}
\usepackage{subcaption}
\usepackage{amsfonts}
\usepackage{mathtools}
\usepackage{balance}

\begin{document}

% \copyrightyear{2018}
% \acmYear{2018}
% \setcopyright{acmcopyright}
% \acmConference[KDD '18]{The 24th ACM SIGKDD International Conference on Knowledge Discovery \& Data Mining}{August 19--23, 2018}{London, United Kingdom}
% \acmBooktitle{KDD '18: The 24th ACM SIGKDD International Conference on Knowledge Discovery \& Data Mining, August 19--23, 2018, London, United Kingdom}
% \acmPrice{15.00}
% \acmDOI{10.1145/3219819.3220048}
% \acmISBN{978-1-4503-5552-0/18/08}

\title{Multi-Cast Attention Networks for Retrieval-based Question Answering and Response Prediction}

\author{Yi Tay}
\affiliation{%
  \institution{Nanyang Technological University Singapore}
  % \streetaddress{P.O. Box 1212}
}
\email{ytay017@e.ntu.edu.sg}

\author{Luu Anh Tuan}
\affiliation{%
 \institution{Institute for Infocomm Research Singapore}
  % \streetaddress{P.O. Box 1212}
}
\email{at.luu@i2r.a-star.edu.sg}

\author{Siu Cheung Hui}
\affiliation{%
  \institution{Nanyang Technological University Singapore}
  % \streetaddress{P.O. Box 1212}
}
\email{asschui@ntu.edu.sg}

\begin{abstract}
Attention is typically used to select informative sub-phrases that are
used for prediction. This paper investigates the novel use of attention
as a form of feature augmentation, i.e, \textit{casted attention}.
We propose Multi-Cast Attention Networks (MCAN), a new attention mechanism
and general model architecture for a potpourri of ranking tasks in the conversational
modeling and question answering domains. Our approach performs a series of soft attention
operations, each time casting a scalar feature upon the inner word embeddings. The key idea
is to provide a real-valued hint (feature) to a subsequent encoder layer and is targeted at improving
the representation learning process. There are several advantages
to this design, e.g., it allows an arbitrary number of attention mechanisms to be casted, allowing for
multiple attention types (e.g., co-attention, intra-attention) and attention variants (e.g., alignment-pooling, max-pooling,
mean-pooling) to be executed simultaneously. This not only eliminates the costly need to tune the nature of the co-attention layer, but also provides greater extents of explainability to practitioners. Via extensive experiments on four well-known benchmark datasets,
we show that MCAN achieves state-of-the-art performance. On the Ubuntu Dialogue Corpus, MCAN outperforms existing state-of-the-art models by $9\%$. MCAN also achieves the best performing score to date on the well-studied TrecQA dataset.
\end{abstract}

\keywords{Deep Learning; Information Retrieval; Question Answering; Conversation Modeling}

\maketitle

\section{Introduction}
Modeling textual relevance between document query pairs lives at the heart of information retrieval (IR) research. Intuitively, this enables
a wide assortment of real life applications, ranging from standard web search to automated chatbots. The key idea is that these systems learn
a scoring function between document-query pairs, providing a ranked list of candidates as an output. A considerable fraction of such IR systems
are focused on short textual documents, e.g., answering facts based questions or selecting the best response in the context of a chat-based system.
The application of retrieval-based response and question answering (QA) systems is overall versatile, potentially serving as a powerful standalone domain-specific system or a crucial component in larger, general purpose chat systems such as Alexa. This paper presents a universal neural ranking model for such tasks.

Neural networks (or deep learning) have garnered considerable attention for retrieval-based systems \cite{Severyn2015,DBLP:conf/acl/WangL016,shen2014latent,DBLP:conf/cikm/YangAGC16,DBLP:conf/naacl/HeL16}. Notably, the dominant state-of-the-art systems for many benchmarks are now neural models, almost completely dispensing with traditional feature engineering techniques altogether. In these systems, convolutional or recurrent networks are empowered with recent techniques such as neural attention \cite{DBLP:conf/acl/WangL016,rocktaschel2015reasoning,bahdanau2014neural}, achieving very competitive results on standard benchmarks. The key idea of attention is to extract only the most relevant information that is useful for prediction. In the context of textual data, attention learns to weight words and sub-phrases within documents based on how important they are. In the same vein, co-attention mechanisms \cite{DBLP:journals/corr/XiongZS16,DBLP:journals/corr/SantosTXZ16,DBLP:conf/aaai/ZhangLSW17,DBLP:conf/emnlp/ShenYD17} are a form of attention mechanisms that learn joint pairwise attentions, with respect to both document and query.

Attention is traditionally used and commonly imagined as a feature extractor. It's behavior can be thought of as a dynamic form of pooling as it learns to select and compose different words to form the final document representation. This paper re-imagines attention as a form of feature augmentation method. Attention is casted with the purpose of not compositional learning or pooling but to provide hints for subsequent layers. To the best of our knowledge, this is a new way to exploit attention in neural ranking models. We begin by describing not only its advantages but also how it handles the weaknesses of existing models designed today.

Typically, attention is applied once to a sentence \cite{rocktaschel2015reasoning,DBLP:conf/acl/WangL016}. A final representation is learned, and then passed to prediction layers. In the context of handling sequence pairs, co-attention is applied and a final representation for each sentence is learned \cite{DBLP:journals/corr/SantosTXZ16,DBLP:conf/aaai/ZhangLSW17,DBLP:conf/emnlp/ShenYD17}. An obvious drawback which applies to many existing models is that they are generally restricted to one attention variant. In the case where one or more attention calls are used (e.g., co-attention and intra-attention, etc.), concatenation is generally used to fuse representations \cite{DBLP:conf/emnlp/ShenYD17,DBLP:conf/emnlp/ParikhT0U16}. Unfortunately, this incurs cost in subsequent layers by doubling the representation size per call.

The rationale for desiring more than one attention call is intuitive. In \cite{DBLP:conf/emnlp/ShenYD17,DBLP:conf/emnlp/ParikhT0U16}, Co-Attention and Intra-Attention are both used because each provides a different view of the document pair, learning high quality representations that could be used for prediction. Hence, this can significantly improve performance. Moreover, Co-Attention also comes in different flavors and can either be used with extractive max-mean pooling \cite{DBLP:journals/corr/SantosTXZ16,DBLP:conf/aaai/ZhangLSW17} or alignment-based pooling \cite{DBLP:conf/emnlp/ParikhT0U16,DBLP:conf/acl/ChenZLWJI17,DBLP:conf/emnlp/ShenYD17}. Each co-attention type produces different document representations. In max-pooling, signals are extracted based on a word's \textit{largest} contribution to the other text sequence. Mean-pooling calculates its contribution to the overall sentence. Alignment-pooling is another flavor of co-attention, which aligns semantically similar sub-phrases together. As such, different pooling operators provide a different view of sentence pairs. This is often tuned as a hyperparameter, i.e., performing architectural engineering to find the best variation that works best on each problem domain and dataset.

Our approach is targeted at serving two important purposes - (1) It removes the need for architectural engineering of this component by enabling attention to be called for an arbitrary $k$ times with hardly any consequence and (2) concurrently it improves performance by modeling multiple views via multiple attention calls.  As such, our method is in similar spirit to multi-headed attention, albeit efficient. To this end, we introduce \textit{Multi-Cast Attention Networks} (MCAN), a new deep learning architecture for a potpourri of tasks in the question answering and conversation modeling domains. In our approach, attention is \textbf{casted}, in contrast to the most other works that use it as a pooling operation. We cast co-attention multiple times, each time returning a compressed \textbf{scalar} feature that is re-attached to the original word representations. The key intuition is that compression enables scalable casting of multiple attention calls, aiming to provide subsequent layers with a \textit{hint} of not only global knowledge but also cross sentence knowledge. Intuitively, when passing these enhanced embeddings into a compositional encoder (such as a long short-term memory encoder), the LSTM can then benefit from this hint and alter its representation learning process accordingly.
\subsection{Our Contributions}
 In summary, the prime contributions of this work are:
\begin{itemize}
\item For the first time, we propose a new paradigm of utilizing attentions not as a pooling operator but as a form of feature augmentation. We propose an overall architecture, Multi-Cast Attention Networks (MCAN) for generic sequence pair modeling.
\item We evaluate our proposed model on four benchmark tasks, i.e., Dialogue Reply Prediction (Ubuntu dialogue corpus), Factoid Question Answering (TrecQA), Community Question Answering (QatarLiving forums from SemEval 2016) and Tweet Reply Prediction (Customer support). On Ubuntu dialogue corpus, MCAN outperforms the existing state-of-the-art models by $9\%$. MCAN also achieves the best performing score of $0.838$ MAP and $0.904$ MRR on the well-studied TrecQA dataset.
\item We provide a comprehensive and in-depth analysis of the inner workings of our proposed MCAN model. We show that the casted attention features are interpretable and are capable of learning (1) a neural adaptation of word overlap and (2) a differentiation of evidence and anti-evidence words/patterns.
\end{itemize}

\section{Multi-Cast Attention Networks}
\begin{figure}[ht]
\centering
  \includegraphics[width=1.0\linewidth]{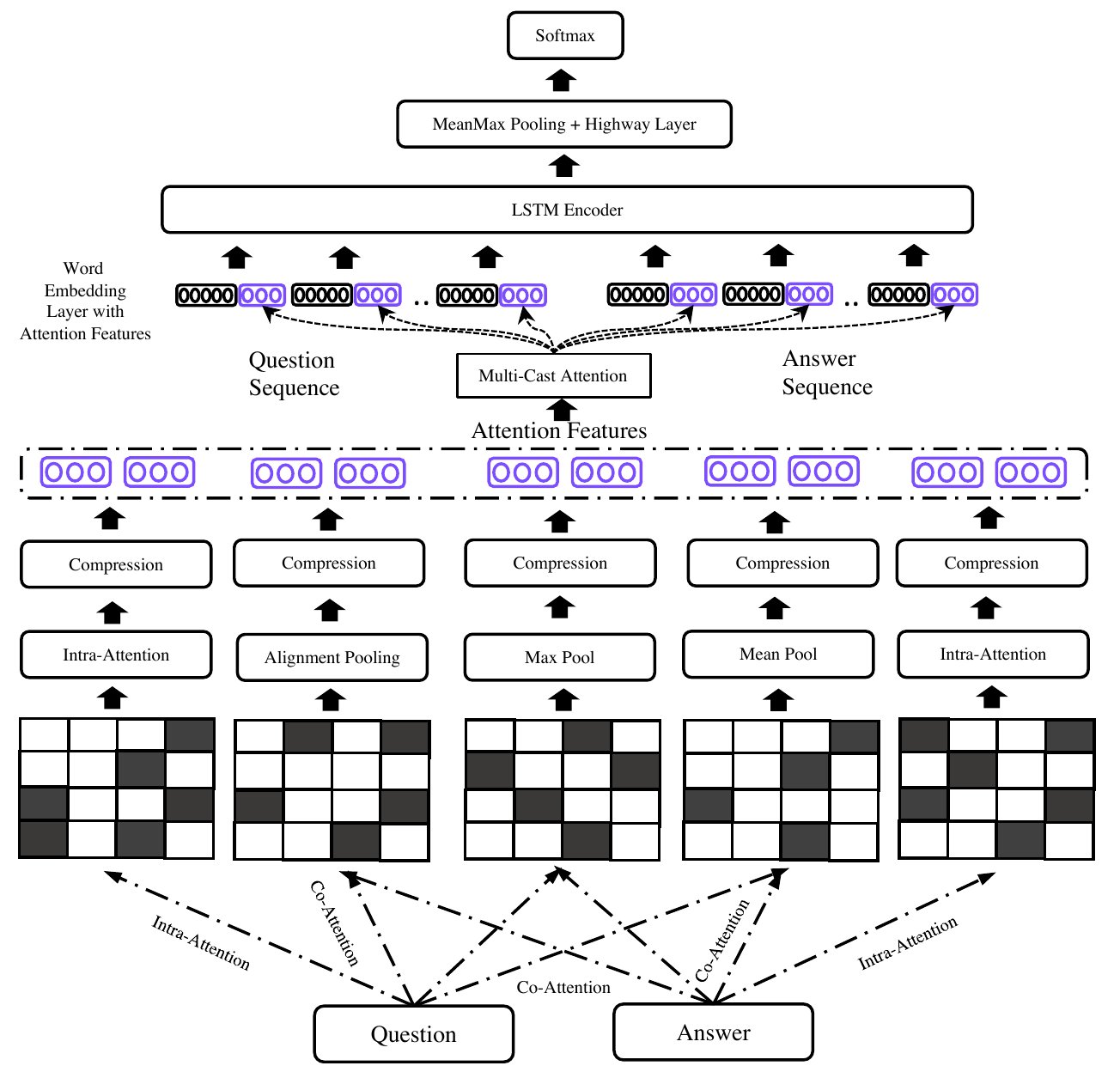}
  \caption{Illustration of our proposed Multi-Cast Attention Networks (\textit{Best viewed in color}). MCAN is a wide multi-headed attention architecture that utilizes compression functions and attention as features. Example is given for question-answer retrieval. Input Encoding layer is ommitted for clarity.}
  \label{fig:architecture}
\end{figure}

In this section, we describe our proposed MCAN model. The inputs to our model are two text sequences which we
denote as query $q$ and document $d$. In our problem, query-document can be generalizable to different problem domains such as question-answering or message-response prediction. Figure \ref{fig:architecture} illustrates the overall model architecture for question-answer retrieval.

\subsection{Input Encoder}
The document and query inputs are passed in as one-hot encoded vectors. A word embedding layer parameterized
by $W_{e} \in \mathbb{R}^{d \times |V|}$ converts each word to a dense word representation $w \in \mathbb{R}^{d}$. $V$ is the set of all
words in the vocabulary.

\subsubsection{Highway Encoder}
Each word embedding is passed through a highway encoder layer. Highway networks \cite{DBLP:journals/corr/SrivastavaGS15}
are gated nonlinear transform layers which control information flow to subsequent layers. Many works adopt a
 projection layer that is trained in place of the raw word embeddings. Not only does this save computation cost but also
 reduces the number of trainable parameters. Our work extends this projection layer to use a highway encoder. The intuition for doing so
 is simple, i.e., highway encoders can be interpreted as data-driven word filters. As such, we can imagine them to parametrically learn which words
 have an inclination to be important and not important to the task at hand. For example, filtering stop words and words that usually do not contribute much to the prediction. Similar to recurrent models that are gated in nature, this highway encoder layer controls how much information (of each word) is flowed to the subsequent layers.

 Let $H(.)$ and $T(.)$ be single layered affine transforms with ReLU and sigmoid activation functions respectively. A single highway network layer is defined as:
 \begin{align}
y = H(x, W_{H}) \cdot T(x, W_{T}) + (1-T(x, W_{T})) \cdot x
 \end{align}
 where $W_H, W_{T} \in \mathbb{R}^{r \times d}$. Notably, the dimensions of the affine transform might be different from the size of the input vector. In this case, an additional nonlinear transform is used to project $x$ to the same dimensionality.

\subsection{Co-Attention}
Co-Attention \cite{DBLP:journals/corr/XiongZS16} is a pairwise attention mechanism that enables attending to text sequence pairs jointly. In this section, we introduce four variants of attention, i.e., (1) max-pooling, (2) mean-pooling, (3) alignment-pooling, and finally (4) intra-attention (or self attention). The first step in co-attention is to learn an affinity (or similarity) matrix between each word across both sequences. Following Parikh et al. \cite{DBLP:conf/emnlp/ParikhT0U16}, we adopt the following formulation for learning the affinity matrix.
\begin{align}
s_{ij} = F(q_{i})^{T} F(d_{j})
\label{aff}
\end{align}
where $F(.)$ is a function such as a multi-layered perceptron (MLP). Alternate forms of co-attention are also possible such as $s_{ij} = q_{i}^{\top} M d_j$ and $s_{ij} = F([q_i;d_j])$.
\subsubsection{Extractive Pooling}
The most common variant of extractive co-attention is the \textit{max-pooling} co-attention, which attends to each word based on its maximum influence it has on the other text sequence.
\begin{align}
q' = Soft(\max_{col}(s))^{\top} q \:\: \: \text{and} \:\: \: d' = Soft(\max_{row}(s))^{\top} d
\end{align}
where $q', d'$ are the co-attentive representations of $q$ and $d$ respectively. Soft(.) is the Softmax operator. Alternatively, the mean row and column-wise pooling of matrix $s$ can be also used:
\begin{align}
q' = Soft(\mathop{mean}_{col}(s))^{\top} q \:\: \: \text{and} \:\: \: d' = Soft(\mathop{mean}_{row}(s))^{\top} d
\end{align}
However, each pooling operator has different impacts and can be intuitively understood as follows: max-pooling selects each word based on its maximum importance of all words in the other text. Mean-pooling is a more \textit{wholesome} comparison, paying attention to a word based on its overall influence on the other text. This is usually dataset-dependent, regarded as a hyperparameter and is tuned to see which performs best on the held out set.

\subsubsection{Alignment-Pooling}
Soft alignment-based pooling has also been utilized for learning co-attentive representations \cite{DBLP:conf/emnlp/ParikhT0U16}. However, the key difference with soft alignment is that it \textit{realigns} sequence pairs while standard co-attention simply learns to weight and score important words. The co-attentive representations are then learned as follows:
\begin{align}
d'_i := \sum^{\ell_{q}}_{j=1} \frac{exp(s_{ij})}{\sum_{k=1}^{\ell_{q}} exp(s_{ik})} q_{j} \:\:\:\text{and} \:\:\:
q'_j := \sum^{\ell_{d}}_{i=1} \frac{exp(s_{ij})}{\sum_{k=1}^{\ell_{d}} exp(s_{kj})} d_{i}
\end{align}
where $d'_i$ is the sub-phrase in $q$ that is softly aligned to $d_i$.
Intuitively, $d'_i$ is a weighted sum across $\{q_j\}^{\ell_{q}}_{j=1}$, selecting the most relevant parts of $q$ to represent $d_i$.

\subsubsection{Intra-Attention}
Intra-Attention, or Self-Attention was recently proposed to learn representations that are aware of long-term dependencies. This is often formulated as an co-attention (or alignment) operation with respect to itself. In this case, we apply intra-attention to both document and query independently. For notational simplicity, we refer to them as $x$ instead of $q$ or $d$ here. The Intra-Attention function is defined as:
\begin{align}
x^{\prime}_i &:= \sum^{\ell}_{j=1} \frac{exp(s_{ij})}{\sum_{k=1}^{\ell} exp(s_{ik})} x_{j}
\end{align}
where $x^{\prime}_i$ is the intra-attentional representation of $x_j$.

\subsection{Multi-Cast Attention}
At this point, it is easy to make several observations. Firstly, each attention mechanism provides a different flavor to the model. Secondly, attention is used to alter the original representation either by re-weighting or realigning. As such, most neural architectures only make use of one type of co-attention or alignment function \cite{DBLP:conf/emnlp/ParikhT0U16,DBLP:journals/corr/SantosTXZ16}. However, this requires the right model architecture to be tuned and potentially missing out from the benefits brought by using multiple variations of co-attention mechanism. As such, our work casts each attention operation as a \textbf{word-level} feature.

\subsubsection{Casted Attention}

Let $x$ be either $q$ or $d$ and $\bar{x}$ is the representation\footnote{We omit subscripts for clarity.} of $x$ after applying co-attention or soft attention alignment. The attention features for the co-attention operators are:
\begin{align}
f_{c} &= F_{c}([\bar{x}; x]) \\
f_{m} &= F_{c}(\bar{x} \odot x) \\
f_{s} &= F_{c}(\bar{x} - x)
\end{align}
where $\odot$ is the Hadamard product and $[.;.]$ is the concatenation operator. $F_{c}(.)$ is a compression function used to reduce features to a scalar. Intuitively, what is achieved here is that we are modeling the influence of co-attention by comparing representations before and after co-attention. For soft-attention alignment, a critical note here is that $x$ and $\bar{x}$ (though of equal lengths) have \textit{`exchanged'} semantics. In other words, in the case of $q$, $\bar{q}$ actually contains the aligned representation of $d$. Finally, the usage of multiple comparison operators (subtractive, concatenation and multiplicative operators) is to capture multiple perspectives and is inspired by the ESIM model \cite{DBLP:conf/acl/ChenZLWJI17}.

\subsubsection{Compression Function}
This section defines $F_{c}(.)$ the compression function used. The rationale for compression is simple and intuitive - we do not want to \textit{bloat} subsequent layers with a high dimensional vector which consequently incurs parameter costs in subsequent layers. We investigate the usage of three compression functions, which are capable of reducing a $n$ dimensional vector to a scalar.
\begin{itemize}
\item \textbf{Sum} (SM) Function is a non-parameterized function that sums the entire vector, returning a scalar as an output.
\begin{align}
F(x) = \sum^{n}_{i}x_{i} \:\:,\:\: \forall x_i \in x
\end{align}
\item \textbf{Neural Network} (NN) is a fully-connected layer that converts each $n$ dimensional feature vector as follows:
\begin{align}
F(x) = ReLU(W_c(x) + b_c).
\end{align}
where $W_c \times \mathbb{R}^{n \times 1}$ and $b_c \in \mathbb{R}$ are the parameters of the FC layer.
\item \textbf{Factorization Machines} (FM) are general purpose machine learning techniques that accept a real-valued feature vector $x \in \mathbb{R}^{n}$ and return a scalar output.
\begin{align}
F(x) &= w_{0} + \sum^{n}_{i=1} w_i \: x_i  +  \sum^{n}_{i=1} \sum^{n}_{j=i+1} \langle v_i, v_j \rangle \: x_i \: x_j
\end{align}
where $w_{0} \in \mathbb{R}\:, w_i \in \mathbb{R}^{n} \: \text{and} \: \{v_1, \cdots v_{n}\} \in \mathbb{R}^{n \times k}$ are the parameters of the FM model. FMs are expressive models that capture pairwise interactions between features using factorized parameters. $k$ is the number of factors of the FM model. For more details, we refer interested readers to \cite{rendle2010factorization}.
\end{itemize}
Note that we do not share parameters across multiple attention casts because each attention cast is aimed at modeling a different view. Our experiments report the above mentioned variants under the model name MCAN (SM), MCAN (NN) and MCAN (FM) respectively.

\subsubsection{Multi-Cast}
The key idea behind our architecture is the facilitation of $k$ attention calls (or casts), with each cast
augmenting raw word embeddings with a real-valued attentional hint. We formally describe the Multi-cast Attention mechanism. For each
query-document pair, we apply (1) Co-Attention with mean-pooling (2) Co-Attention with max-Pooling and (3) Co-Attention with alignment-pooling. Additionally,
we apply Intra-Attention to both query and document individually. Each attention cast produces three scalars (per word) which are concatenated with the word embedding. The final casted feature vector is $z \in \mathbb{R}^{12}$. As such, for each word $w_i$, the new word representation becomes $\bar{w}_i = [w_i; z_i]$.

\subsection{Long Short-Term Memory Encoder}
Next, the word representations with casted attention $\bar{w}_1, \bar{w}_2, \dots \bar{w}_{\ell}$ are then passed into a sequential encoder layer. We adopt a standard vanilla long short-term memory (LSTM) encoder:
\begin{align}
h_i = LSTM(u, i), \forall i \in [1, \dots \ell]
\end{align}
where $\ell$ represents the maximum length of the sequence. Notably, the parameters of the LSTM are \textit{`siamese'} in nature, sharing weights between document and query. The key idea is that the LSTM encoder learns representations that are aware of sequential dependencies by the usage of nonlinear transformations as gating functions. Since LSTMs are standard neural building blocks, we omit technical details in favor of brevity. As such, the key idea behind casting attention as features right before this layer is that it provides the LSTM encoder with hints that provide information such as (1) long-term and global sentence knowledge and (2) knowledge between sentence pairs (document and query).

\subsubsection{Pooling Operation}
 Finally, a pooling function is applied across the hidden states $\{h_1 \dots h_{\ell}\}$ of each sentence, converting the sequence into a fixed dimensional representation.
 \begin{align}
h = \mathop{MeanMax} [h_1 \dots h_{\ell}]
\end{align}
 We adopt the $\mathop{MeanMax}$ pooling operator, which concatenates the result of the mean pooling and max pooling together. We found this to consistently perform better than using $\mathop{max}$ or $\mathop{mean}$ pooling in isolation.

\subsection{Prediction Layer and Optimization}

Finally, given a fixed dimensional representation of the document-query pair, we pass their concatenation into a two-layer $h$-dimensional highway network. The final prediction layer of our model is computed as follows:
\begin{align}
y_{out} = H_{2}(H_{1}([x_{q}; x_{d}; x_q \odot x_d; x_q - x_d]))
\end{align}
where $H_{1}(.), H_{2}(.)$ are highway network layers with ReLU activation. The output is then passed into a final linear softmax layer.
\begin{align}
y_{pred} = softmax(W_{F} \cdot y_{out} + b_{F})
\end{align}
where $W_{F} \in \mathbb{R}^{h \times 2}$ and $b_{F} \in \mathbb{R}^{2}$. The network is then trained using standard multi-class cross entropy loss with L2 regularization.

\begin{align}
J(\theta) = -\sum^{N}_{i=1} \: [ y_i \log \hat{y}_i + (1-y_i)\log(1-\hat{y}_i)] + \lambda||\theta||_{L2}
\end{align}
where $\theta$ are the parameters of the network. $\hat{y}$ is the output of the network. $||\theta||_{L2}$ is the L2 regularization and $\lambda$ is the weight of the regularizer.

\section{Empirical Evaluation}
In our experiments, we aim to answer the following research questions (\textbf{RQs}):
\begin{enumerate}
\item \textbf{RQ1} - Does our proposed approach achieve state-of-the-art performance on question answering and conversation modeling tasks? What is the relative improvement over well-established baselines?
\item \textbf{RQ2} - What are the impacts of architectural design on performance? Is the LSTM encoder necessary to make use of the casted features? Does all the variations of co-attention contribute to the overall model performance?
\item \textbf{RQ3} - Can we explain the inner workings of our proposed model? Can we interpret the casted attention features?
\end{enumerate}

\subsection{Experiment 1 - Dialogue Prediction}
In this first task, we evaluate our model on its ability to successfully predict the next reply in conversations.

\subsubsection{Dataset and Evaluation Metric}
For this experiment, we utilize the large and well-known large-scale Ubuntu Dialogue Corpus (UDC) \cite{lowe2015ubuntu}. We use the same
testing splits provided by Xu et al. \cite{xu2016incorporating}. In this task, the goal is to match a sentence with its reply.
Following \cite{wu2016knowledge}, the task mainly utilizes the last two utterances in each conversation, predicting if the latter
follows the former. The training set consists of \textbf{one million} message-response pairs with a $1:1$ positive-negative ratio.
The development and testing sets have a $9:1$ ratio. Following \cite{wu2016knowledge,xu2016incorporating}, we use the evaluation metrics of
recall$@k$ ($R_n @K$) which indicates whether the ground truth exists in the top $k$ results from $n$ candidates. The four evaluation metrics used are $R_{2}@1$, $R_{10}@1$, $R_{10}@2$ and $R_{10}@5$.

\subsubsection{Competitive Baselines and Implementation Details}
We compare against a large number of competitive baselines, e.g., MLP, DeepMatch \cite{lu2013deep}, ARC-I / ARC-II \cite{Hu2014a},
CNTN \cite{DBLP:conf/ijcai/QiuH15}, MatchPyramid \cite{pang2016text}, vanilla LSTM, Attentive Pooling LSTM \cite{DBLP:journals/corr/SantosTXZ16}, MV-LSTM \cite{DBLP:conf/aaai/WanLGXPC16} and finally the state-of-the-art Knowledge Enhanced Hybrid Neural Network (KEHNN) \cite{wu2016knowledge}. A detailed description of baselines can be found at \cite{wu2016knowledge}. Since testing splits are the same, we report
the results directly from \cite{wu2016knowledge}. For fair comparison, we set the LSTM encoder size in MCAN to $d=100$ which makes it equal to the models in \cite{wu2016knowledge}. We optimize MCAN with Adam optimizer \cite{DBLP:journals/corr/KingmaB14} with an initial learning rate of $3 \times 10^{-4}$. A dropout rate of $0.2$ is applied to all layers except the word embedding layer. The sequences are dynamically truncated or padded to their batch-wise maximums (with a hard limit of $50$ tokens). We initialize the word embedding layer with pretrained GloVe embeddings.

\subsubsection{Experimental Results}
Table \ref{tab:ubuntu} reports the results of our experiments. Clearly, we observe that all MCAN models achieve a huge performance gain over existing state-of-the-art models. More specifically, the improvement across all metrics are $\approx 5\%-9\%$ better than KEHNN. The performance improvement over strong baselines such as AP-LSTM and MV-LSTM are even greater, hitting an improvement of $15\%$ in terms of $R_{10}@{1}$. This ascertains the effectiveness of the MCAN model. Overall, MCAN (FM) and MCAN (NN) are comparable in terms of performance. MCAN (SM) is marginally lower than both MCAN (FM) and MCAN (NN). However, its performance is still considerably higher than the existing state-of-the-art models.
% Table generated by Excel2LaTeX from sheet 'Sheet2'
\begin{table}[H]
  \centering

    \begin{tabular}{lrrrr}
    \hline
          & \multicolumn{1}{l}{$R_2@1$} & \multicolumn{1}{l}{$R_{10}@1$} & \multicolumn{1}{l}{$R_{10}@2$} & \multicolumn{1}{l}{$R_{10}@5$} \\
          \hline
    MLP   & 0.651 & 0.256 & 0.380  & 0.703 \\
    DeepMatch & 0.593 & 0.345 & 0.376 & 0.693 \\
    ARC-I & 0.665 & 0.221 & 0.360  & 0.684 \\
    ARC-II & 0.736 & 0.380  & 0.534 & 0.777 \\
    CNTN  & 0.743 & 0.349 & 0.512 & 0.797 \\
    MatchPyramid & 0.743 & 0.420  & 0.554 & 0.786 \\
    LSTM  & 0.725 & 0.361 & 0.494 & 0.801 \\
    AP-LSTM & 0.758 & 0.381 & 0.545 & 0.801 \\
    MV-LSTM & 0.767 & 0.410  & 0.565 & 0.800 \\
    KEHNN & 0.786 & 0.460  & 0.591 & 0.819 \\
    \hline
    MCAN (SM) & 0.831& 0.548&0.682 &0.873 \\
    MCAN (NN) & \underline{0.833} & \underline{0.549} & \textbf{0.686}
     & \textbf{0.875} \\
    MCAN (FM) & \textbf{0.834} & \textbf{0.551} & \underline{0.684} & \textbf{0.875} \\
    \hline
    \end{tabular}%
     \caption{Performance Comparison on Ubuntu Dialogue Corpus. Best result is in boldface and second best is underlined. }
  \label{tab:ubuntu}%
\end{table}%

\subsection{Experiment 2 - Factoid Question Answering}
Factoid question answering is the task of answering factual based questions. In this task, the goal is to
provide a ranked list of answers to a given question.

\subsubsection{Dataset and Evaluation Metric}
We utilize the QA dataset from TREC (Text Retrieval Conference). TrecQA is
one of the most widely evaluated dataset, competitive and long standing benchmark for QA. This dataset was prepared by Wang et al. \cite{Wang2007} and contains
$53K$ QA pairs for training and $1100/1500$ pairs for development and testing respectively. Following the recent works, we evaluate on the clean setting as noted by \cite{DBLP:conf/cikm/RaoHL16}. The evaluation metrics for this task are the MAP (mean average precision) and MRR (mean reciprocal rank) scores which are well-known IR metrics.

\subsubsection{Competitive Baselines and Implementation Details}
 We compare against all previously published works on this dataset. The competitive baselines for this task are QA-LSTM / AP-CNN \cite{DBLP:journals/corr/SantosTXZ16}, LDC model \cite{wang2016sentence}, MPCNN \cite{DBLP:conf/emnlp/HeGL15}, MPCNN+NCE \cite{DBLP:conf/cikm/RaoHL16}, HyperQA \cite{Tay:2018:HRL:3159652.3159664}, BiMPM \cite{DBLP:conf/ijcai/WangHF17} and IWAN \cite{DBLP:conf/emnlp/ShenYD17}. For our model, the size of the LSTM used is $300$. The dimensions of the highway prediction layer is $200$. We use the Adam optimizer with a $3 \times 10^{-4}$ learning rate. The L2 regularization is set to $10^{-6}$. A dropout rate of $0.2$ is applied to all layers except the embedding layer. We use pretrained $300d$ GloVe embeddings and fix the embeddings during training. For MCAN (FM), we use a FM model with $10$ factors. We pad all sequences to the maximum sequence length and truncate them to the batch-wise maximums.
% Table generated by Excel2LaTeX from sheet 'Sheet2'

\subsubsection{Experimental Results}
Table \ref{tab:trec} reports our results on TrecQA. All MCAN variations outperform all existing state-of-the-art models. Notably, MCAN (FM) is currently the best performing model on this extensively studied dataset. MCAN (NN) comes in second which marginally outperforms the highly competitive and recent IWAN model. Finally, MCAN (SM) remains competitive to IWAN, despite naively summing over casted attention features.
\begin{table}[H]
  \centering

    \begin{tabular}{lcc}
    \hline
    Model & \multicolumn{1}{c}{MAP} & \multicolumn{1}{c}{MRR} \\
    \hline
    QA-LSTM (dos Santos et al.) & 0.728 & 0.832\\
    AP-CNN (dos Santos et al.) & 0.753 &0.851 \\
    LDC Model (Wang et al.) & 0.771 & 0.845 \\
    MPCNN (He et al.) &0.777 & 0.836 \\
    HyperQA (Tay et al.) & 0.784 & 0.865 \\
    MPCNN + NCE (Rao et al.) & 0.801 & 0.877 \\
    BiMPM (Wang et al.) & 0.802 & 0.899 \\
    IWAN (Shen et al.) & 0.822 & 0.889 \\
    \hline
    MCAN (SM) & 0.827& 0.880\\
    MCAN (NN) & \underline{0.827}& \underline{0.890}\\
    MCAN (FM) & \textbf{0.838} & \textbf{0.904} \\
    \hline
    \end{tabular}%
      \caption{Performance Comparison on TrecQA (\textit{clean}) dataset. Best result is in boldface and second best is underlined. }
  \label{tab:trec}%
\end{table}%

\subsection{Experiment 3 - Community Question Answering (cQA)}
This task is concerned with ranking answers in community forums. Different from factoid QA, answers are generally subjective instead of factual. Moreover, answer lengths are also much longer.
\subsubsection{Dataset and Evaluation}
We use the QatarLiving dataset, a well-studied benchmark dataset from SemEval-2016 Task 3 Subtask A (cQA) and have been used extensively as a benchmark for recent state-of-the-art neural network models for cQA \cite{DBLP:conf/aaai/ZhangLSW17,1711.07656}. This is a real world dataset obtained from Qatar Living Forums and comprises $36K$ training pairs, $2.4K$ development pairs and $3.6K$ testing pairs. In this dataset, there are ten answers in each question `thread' which are marked as `Good`, `Potentially Useful' or `'Bad'. Following \cite{DBLP:conf/aaai/ZhangLSW17}, `Good' is regarded as positive and anything else is regarded as negative labels. We evaluate on two metrics, namely the Precision@1 (P@1) and Mean Average Precision (MAP) metric.

\subsubsection{Competitive Baselines and Implementation Details}
The key competitors of this dataset are the CNN-based ARC-I/II architecture by Hu et al. \cite{Hu2014a}, the Attentive Pooling CNN \cite{DBLP:journals/corr/SantosTXZ16}, Kelp \cite{DBLP:conf/semeval/FiliceCMB16} a feature engineering based SVM method, ConvKN \cite{DBLP:conf/semeval/Barron-CedenoMJ16} a combination of convolutional tree kernels with CNN and finally  AI-CNN (Attentive Interactive CNN) \cite{DBLP:conf/aaai/ZhangLSW17}, a tensor-based attentive pooling neural model. We also compare with the \textit{Cross Temporal Recurrent Networks} (CTRN) \cite{1711.07656}, a recently proposed model for ranking QA pairs which have achieved very competitive performance on this dataset. Following \cite{1711.07656}, we initialize MCAN with domain-specific 200 dimensional word embeddings using the unannotated QatarLiving corpus. Word embeddings are not updated during training. The size of the highway projection layer, LSTM encoder and highway prediction layer are all set to $200$. The model is optimized with Adam optimizer with learning rate of $3 \times 10^{-4}$.

\subsubsection{Experimental Results}

% Table generated by Excel2LaTeX from sheet 'Sheet2'
\begin{table}[htbp]
  \centering

    \begin{tabular}{lcc}
    \hline
   Model & \multicolumn{1}{c}{P@1} & \multicolumn{1}{c}{MAP} \\
   \hline
    ARC-I (Hu et al.) & 0.741 & 0.771 \\
    ARC-II (Hu et al.) & 0.753 & 0.780 \\
    AP-CNN (dos Santos et al.) & 0.755 & 0.771 \\
    Kelp (Filice et al.) & 0.751 & 0.792 \\
    ConvKN (Barron Cedeno et al.) & 0.755 & 0.777 \\
    AI-CNN (Zhang et al.) & 0.763 & 0.792 \\
    CTRN (Tay et al.)  & 0.788 & \underline{0.794} \\
    \hline
    MCAN (SM) & \underline{0.803}& 0.787 \\
    MCAN (NN) & 0.802& 0.784 \\
    MCAN (FM) & \textbf{0.804} & \textbf{0.803} \\
    \hline
    \end{tabular}%
      \caption{Performance comparison on QatarLiving dataset for community question answering. Best result is in boldface and second best is underlined.}
  \label{tab:cqa_ql}%
\end{table}%

Table \ref{tab:cqa_ql} reports the performance comparison on the QatarLiving dataset. Our best performing MCAN model achieves state-of-the-art performance on this dataset. Performance improvement over recent, competitive neural network baselines is significant. Notably, the improvement of MCAN (FM) over AI-CNN on the $P@1$ metric is $4.1\%$ and $1.1\%$ in terms of MAP. MCAN (FM) also achieves competitive results relative to the CTRN model. The performance of MCAN (NN) and MCAN (SM) is lower than MCAN (FM) but still remains competitive on this benchmark.

\subsection{Experiment 4 - Tweet Reply Prediction}
This experiment is concerned with predicting an appropriate reply given a tweet.
\subsubsection{Dataset and Evaluation Metrics}
We utilize a customer support dataset obtained from Kaggle\footnote{\url{https://www.kaggle.com/soaxelbrooke/customer-support-on-twitter}}. This dataset contains tweet-response pairs of tweets to famous brands and their replies. For each Tweet-Reply pair, we randomly selected \textit{four} tweets as negative samples that originate from the same brand. The dataset is split into $8:1:1$ train-dev-test split. The evaluation metrics for this task are MRR (Mean reciprocal rank) and Precision@1 (accuracy). Unlike previous datasets, there are no published works on this dataset. As such, we implement the baselines ourselves. We implement standard baselines such as (1) CBOW (sum embeddings) passed into a 2 layer MLP with ReLU activations, (2) standard vanilla LSTM and CNN models and (3) BiLSTM and CNN with standard Co-Attention (AP-BiLSTM, AP-CNN) following \cite{DBLP:journals/corr/SantosTXZ16}. All models minimize the binary cross entropy loss (pointwise) since we found performance to be much better than using ranking loss. We also include the recent AI-CNN (Attentive Interactive CNN) which uses multi-dimensional co-attention. We set all LSTM dimensions to $d=100$ and the number of CNN filters is $100$. The CNN filter width is set to $3$. We train all models with Adam optimizer with $3 \times 10^{-4}$ learning rate. Word embeddings are initialized with GloVe and fixed during training. A dropout of $0.2$ is applied to all layers except the word embedding layer.

\subsubsection{Experimental Results}
Table \ref{tab:tweets} reports our results on the Tweets dataset. MCAN (FM) achieves the top performance by a significant margin. The performance of MCAN (NN) falls short of MCAN (FM), but is still highly competitive. Our best MCAN model outperforms AP-BiLSTM by $3.4\%$ in terms of MRR and $5.3\%$ in terms of P@1. The performance improvement of AI-CNN is even greater, i.e., $8.4\%$ in terms of MRR and $12.5\%$ in terms of P@1. The strongest baseline is AP-BiLSTM which significantly outperforms AI-CNN and AP-CNN.
% Table generated by Excel2LaTeX from sheet 'Sheet2'
\begin{table}[htbp]
  \centering
    \begin{tabular}{lcc}
    \hline
          Model & \multicolumn{1}{c}{MRR} & \multicolumn{1}{c}{P@1} \\
          \hline
    CBOW + MLP   & 0.658 & 0.442 \\
    LSTM  & 0.652 & 0.431 \\
    CNN   & 0.657 & 0.441 \\
    AP-CNN (dos Santos et al.) & 0.643 & 0.426 \\
    AI-CNN (Zhang et al.) & 0.675&0.465 \\
    AP-BiLSTM (dos Santos et al.) & 0.725 & 0.540 \\
    % AP-BiLSTM & & \\
    % HyperQA & & \\

    \hline
    MCAN (SM) & 0.722& 0.548\\
    MCAN (NN) & \underline{0.747}& \underline{0.585} \\
    MCAN (FM)  & \textbf{0.759} & \textbf{0.593} \\
    \hline
    \end{tabular}%
    \caption{Performance comparison on Reply Prediction on Tweets dataset. Best performance is in boldface and second best is underlined. }
  \label{tab:tweets}%
\end{table}%

\subsection{Ablation Analysis}
This section aims to demonstrate the relative effectiveness of different components of our proposed MCAN model. Table \ref{ablation} reports the results on the validation set of the TrecQA dataset. We report the scores of seven different configurations. In (1), we replace all highway layers with regular feed-forward neural networks. In (2), we remove the LSTM encoder before the prediction layer. In (3), we remove the entire multi-cast attention mechanism. This is equivalent to removing the twelve attention features. In (4-7), we remove different attention casts, aiming to showcase that removing either one results in some performance drop.

From our ablation analysis, we can easily observe the crucial components to our model. Firstly, we observe that removing MCA entirely significantly decreases the performance (ablation 3). In this case, validation MAP drops from 0.866 to 0.670. As such, our casted attention features contribute a lot to the performance of the model. Secondly, we also observe that the LSTM encoder is necessary. This is intuitive because the goal of MCAN is to provide features as hints for a compositional encoder. As such, removing the LSTM encoder allows our attention hints to go unused. While the upper prediction might still manage to learn from these features, it is still sub-optimal compared to using a LSTM encoder. Thirdly, we observed that removing Max or Mean Co-Attention decreases performance marginally. However, removing the Alignment Co-Attention decreases the performance significantly. As such, it is clear that the alignment-based attention is most important for our model. However, Max, Mean and Intra attention all contribute to the performance of MCAN. Hence, using multiple attention casts can improve performance. Finally, we also note that the highway layers also contribute slightly to performance.

\begin{table}[H]
\begin{tabular}{lcc}
\hline
Setting & MAP & MRR \\
\hline
Original & \textbf{0.866} & \textbf{0.922} \\
(1) Remove Highway & 0.825 & 0.863 \\
(2) Remove LSTM & 0.765 & 0.809 \\
(3) Remove MCA & 0.670 & 0.749 \\
(4) Remove Intra & 0.834 & 0.910 \\
(5) Remove Align & 0.682 & 0.726 \\
(6) Remove Mean & 0.858 & 0.906 \\
(7) Remove Max & 0.862 & 0.915 \\
\hline
\end{tabular}

\caption{Ablation analysis (validation set) on TrecQA dataset. }
\label{ablation}
\end{table}

\subsection{In-depth Model Analysis}
In this section, we aim to provide insights pertaining to the inner workings of our model. More specifically,
we list several observations by visual inspection of the casted attention features. We trained a MCAN model with FM compression and extracted the word-level casted attention features. The features are referred to as $f_i$ where $i \in [1,12]$. $f_1, f_2, f_3$ are generated from alignment-pooling. $f_4, f_5, f_6$ and $f_7,f_8,f_9$ are generated from max and mean pooled co-attention respectively. $f_{10},f_{11},f_{12}$ are generated from intra-attention.

\subsubsection{Observation 1: Features learn a Neural Adaptation of Word Overlap}
\begin{figure}[H]
    \centering
    \begin{subfigure}[t]{0.5\textwidth}
        \centering
        \includegraphics[width=1.0\textwidth]{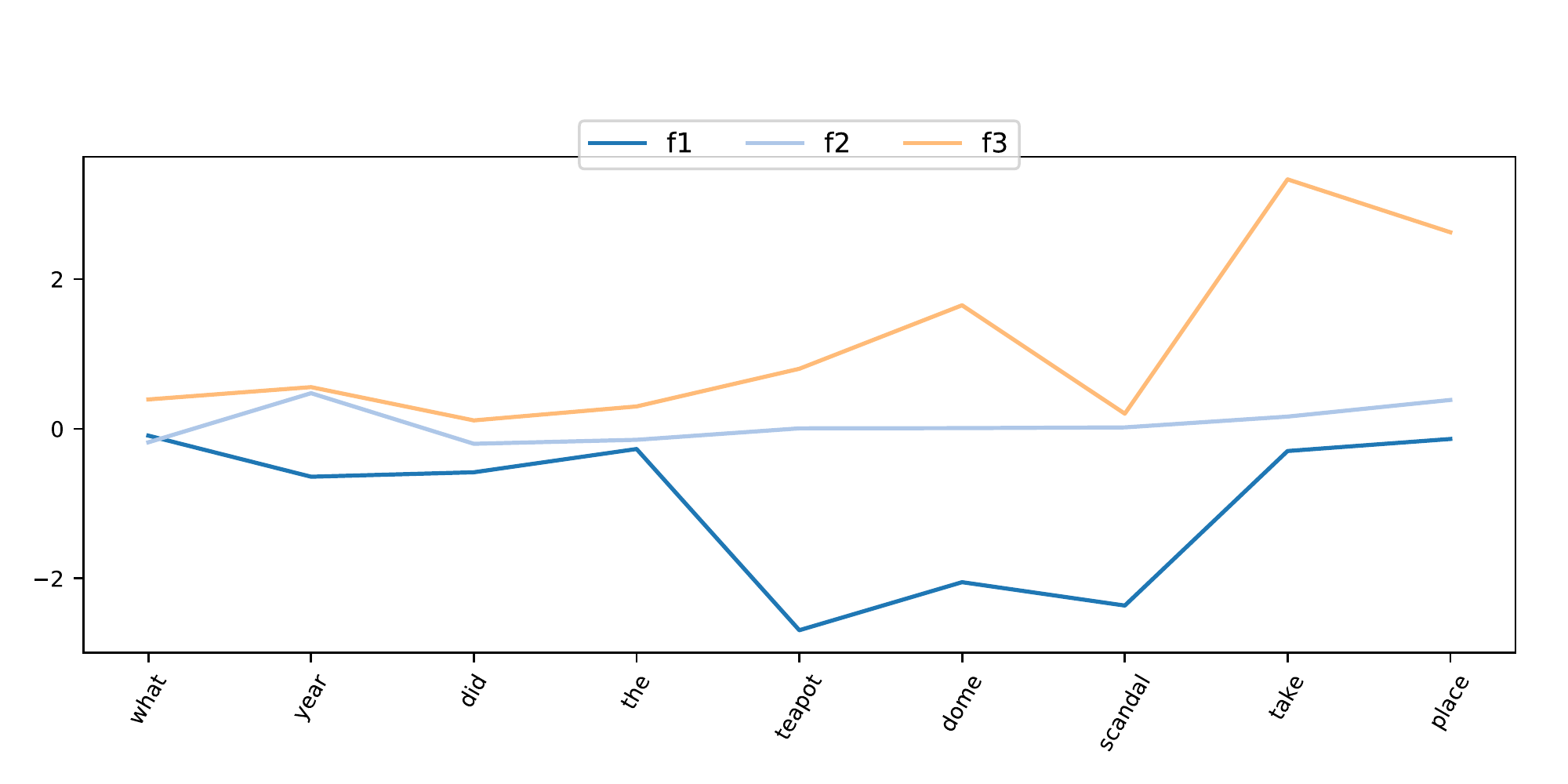}
        \caption{Features $f_1, f_2, f_3$ for question.}
    \end{subfigure}%

    \begin{subfigure}[t]{0.5\textwidth}
        \centering
        \includegraphics[width=1.0\textwidth]{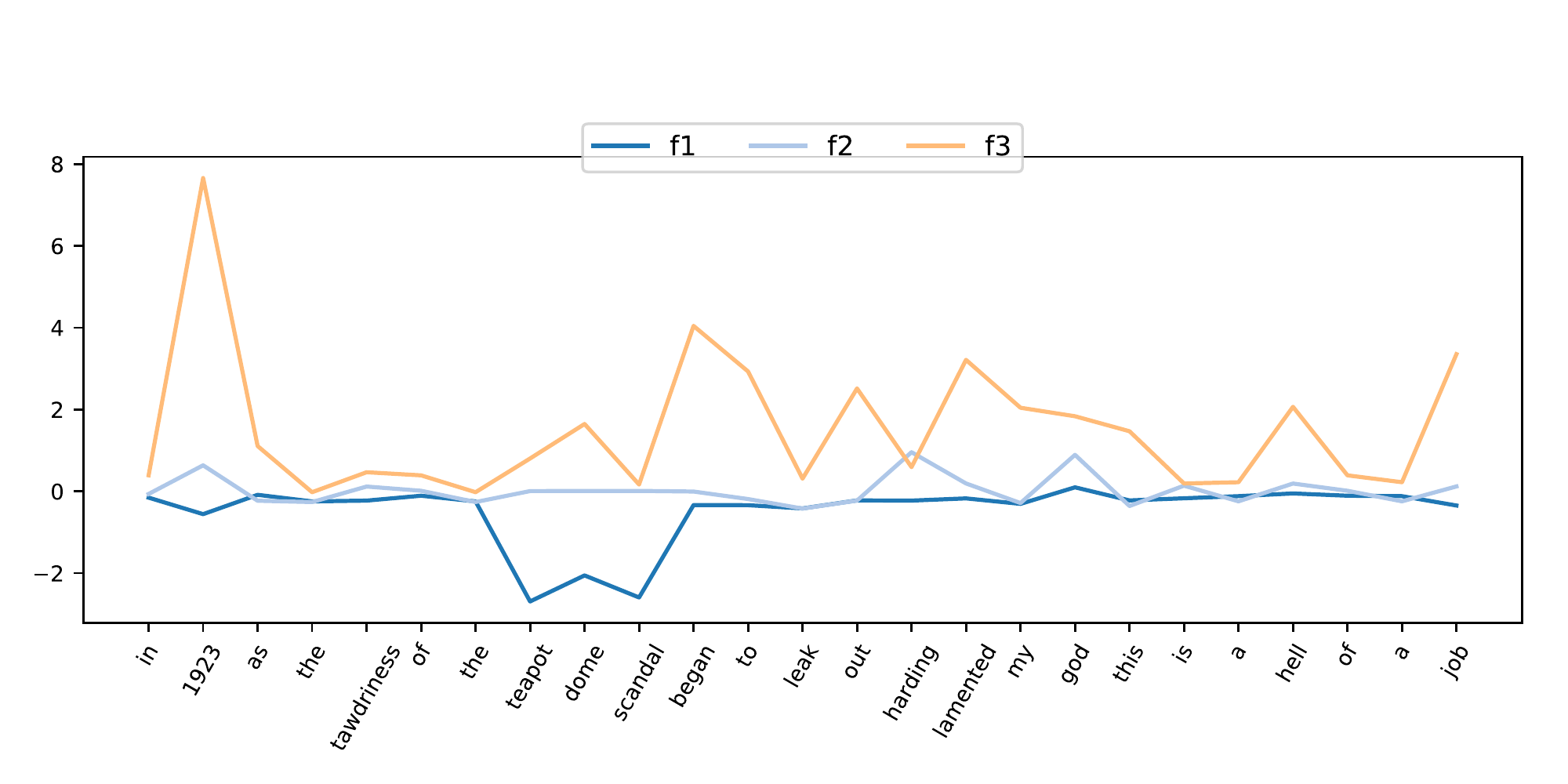}
        \caption{Features $f_1, f_2, f_3$ for answer.}
    \end{subfigure}
    \caption{Visualization of Casted Attention Features ($f_1,f_2,f_3$) on a \textit{positive} test sample from TrecQA. }
\label{intra_pos}
\end{figure}

Figure \ref{intra_pos} and Figure \ref{intra_neg} show a positive and negative QA pair from the TrecQA test set. Firstly,
we analyze\footnote{This is done primarily for clear visualisation, lest the diagram
becomes too cluttered.} the first three features $f_1, \: f_2 \: \text{and} f_3$. These features correspond to the alignment attention
and multiply, subtract and concat composition respectively. From the figures, we observe that $f_1$ spikes (in the negative direction) when there is a word overlap
across sentences, e.g., `\textit{teapot}' in Figure \ref{intra_neg} and \textit{`teapot dome scandal'} in Figure \ref{intra_pos}. Hence, $f_1$ (dark blue line) behaves as a neural adaptation of the conventional overlap feature. Moreover, in contrary to
traditional binary overlap features, we also notice that the value of the neural word overlap feature is dependent on the word itself, i.e., `\textit{teapot}' and `\textit{dome}' have different values. As such, it encodes more information over the traditional binary feature.

\subsubsection{Observation 2: Features React to Evidence and Anti-Evidence}
\begin{figure}
    \centering
    \begin{subfigure}[t]{0.5\textwidth}
        \centering
        \includegraphics[width=1.0\textwidth]{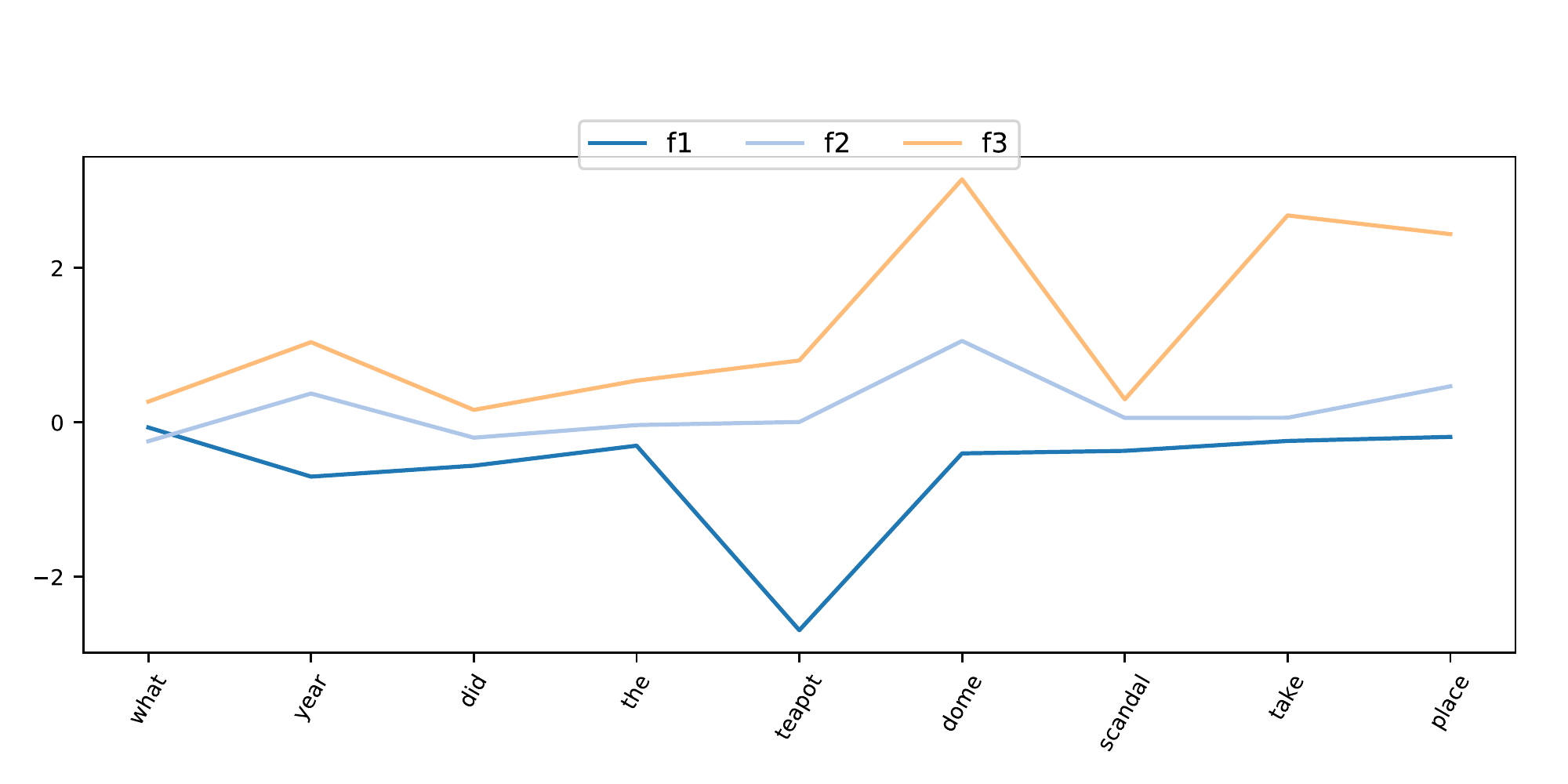}
        \caption{Features $f_1, f_2, f_3$ for question.}
    \end{subfigure}%

    \begin{subfigure}[t]{0.5\textwidth}
        \centering
        \includegraphics[width=1.0\textwidth]{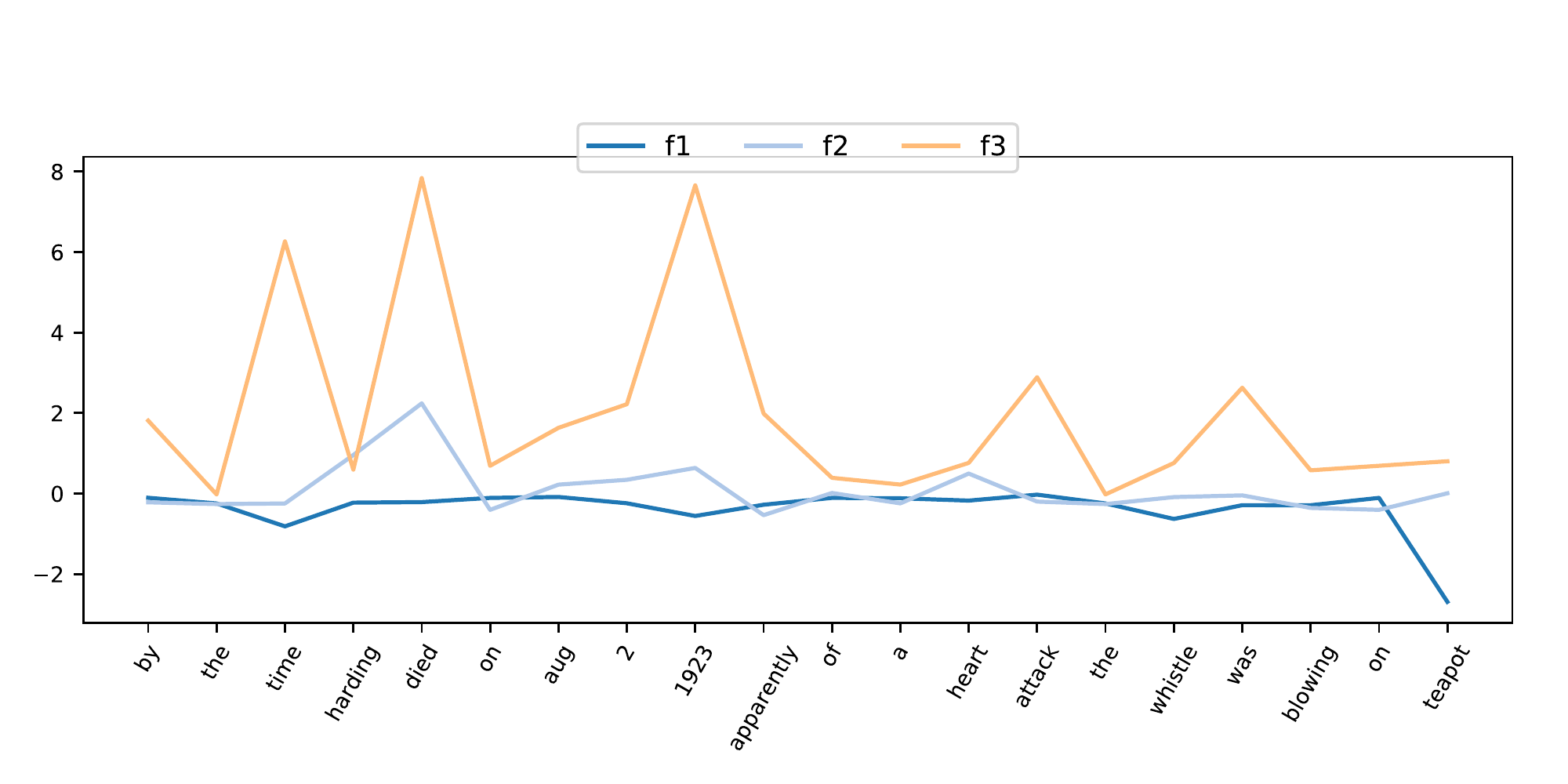}
        \caption{Features $f_1, f_2, f_3$ for answer.}
    \end{subfigure}
     \caption{Visualization of Casted Attention Features ($f_1,f_2,f_3$) on a \textit{negative} test sample from TrecQA. }

\label{intra_neg}
\end{figure}

While $f_1$ is primarily aimed at modeling overlap, we observe that $f_3$ tries to gather supporting evidence for the given QA pair. In Figure \ref{intra_pos}, the words \textit{`year'} and \textit{`1923'} have spiked. It also tries to extract key verbs such as \textit{`'take place'} (question) and \textit{`began'} (answer) which are related verbs generally used to describe events. Finally, we observe that $f_2$ (subtractive composition) seems to be searching for anti-evidence, i.e., a contradictory or irrelevant information. However, this appears to be more subtle as compared to $f_1$ and $f_3$. In Figure \ref{intra_neg}, we note that the words \textit{`died' \text{and} `attack'} (answer) have spiked. We find this example particularly interesting because the correct answer `\textit{1923}' is in fact found in the answer. However, the pair is \textbf{wrong} because the text sample refers to the `\textit{death of Harding}' and does not answer the question correctly. In the negative answer, we found that the word \textit{`died'} has the highest $f_2$ value. As such, we believe that $f_2$ is actively finding anti-evidence to why this QA pair should be negative. Additionally, irrelevant words such as \textit{`attack`} and \textit{`god'} experience nudges by $f_2$. Finally, it is good to note that MCAN classifies these two samples correctly while a standard Bidirectional LSTM does not.

\begin{figure}
    \centering
    \begin{subfigure}[t]{0.5\textwidth}
        \centering
        \includegraphics[width=1.0\textwidth]{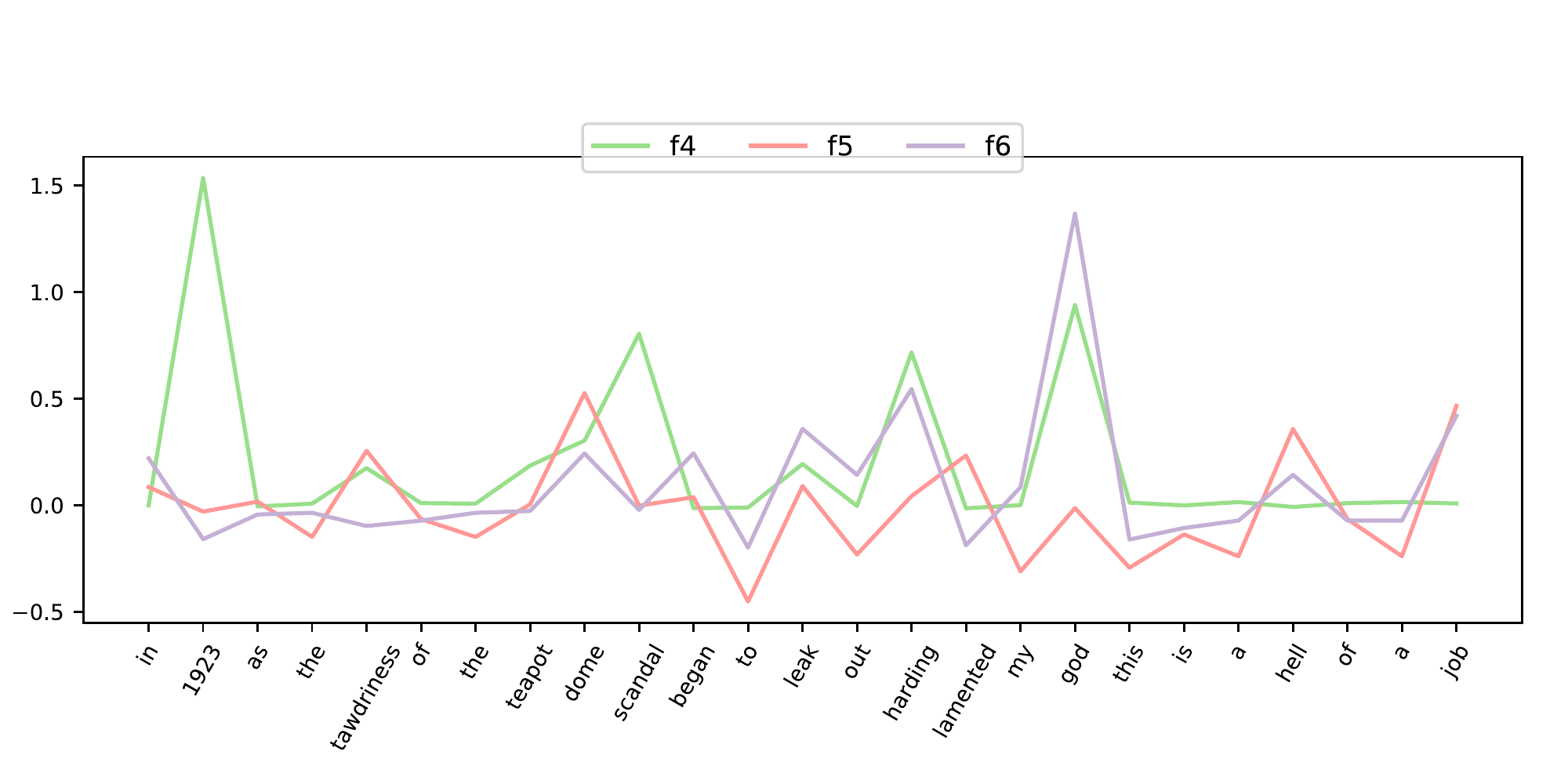}
        \caption{Features generated from \textit{max}-pool Co-Attention.}
    \end{subfigure}%

    \begin{subfigure}[t]{0.5\textwidth}
        \centering
        \includegraphics[width=1.0\textwidth]{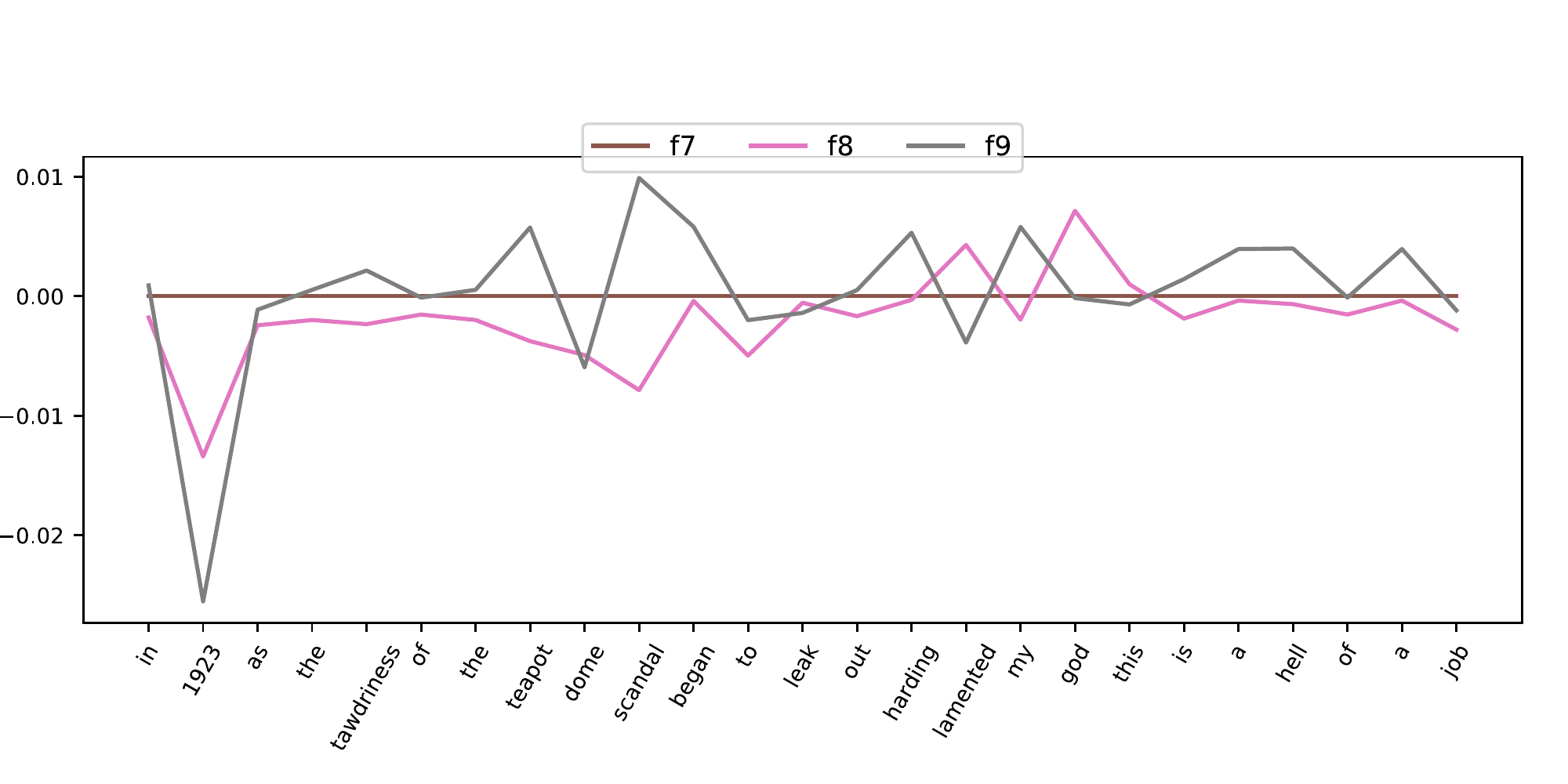}
        \caption{Features generated from \textit{mean}-pool Co-Attention.}
    \end{subfigure}

     \caption{Differences between Max and Mean-pooled Casted Attention Features on answer text from TrecQA dataset. Diverse features
     are learned by different attention casts.}
\label{diversity}
\end{figure}

\subsubsection{Observation 3: Diversity of Multiple Casts}
One of the key motivators for a multi-casted attention is that each attention cast produces features from different views of the sentence pair. While we have shown in our ablation study that all attention casts contributed to the overall performance, this section qualitatively analyzes the output features. Figure \ref{diversity} shows the casted attention features (answer text) for max-pooled attention ($f_4, f_5, f_6$) and mean-pooled\footnote{The values on $f_7$ are not constant. They appear to be since the max-min range is much smaller than $f_8$ and $f_9$.} attention ($f_7, f_8, f_9$). Note that the corresponding question is the same as Figure \ref{intra_pos} and Figure \ref{intra_neg} which allows a direct comparison with the alignment-based attention. We observe that both attention casts produce extremely diverse features. More specifically, not only the spikes are all at different words but the overall sequential pattern is very different. We also note that the feature patterns differ a lot from alignment-based attention (Figure \ref{intra_pos}). While we were aiming to capture more diverse patterns, we also acknowledge that these features are much less interpretable than $f_1,f_2$ and $f_3$. Even so, some patterns can still be interpreted, e.g., the value of $f_5$ is high for important words and low (negative) whenever the words are generic and unimportant such as \textit{`to', `the', `a'}. Nevertheless, the main objective here is to ensure that these features are not learning identical patterns.

\section{Related Work}
Learning to rank short document pairs is a long standing problem in IR research. The dominant state-of-the-art models for learning-to-rank today are mostly neural network based models. Neural network models, such as convolutional neural networks (CNN) \cite{Hu2014a,shen2014latent,DBLP:conf/sigir/TayPLH17,DBLP:conf/acl/WangN15}, recurrent neural networks (RNN) \cite{Mueller2016,DBLP:conf/emnlp/ShenYD17,wu2016knowledge} or recursive neural networks \cite{wan2016match} are used for learning document representations. A parameterized fuction such as multi-layered perceptrons \cite{Severyn2015}, tensor layers \cite{DBLP:conf/ijcai/QiuH15} or holographic layers \cite{DBLP:conf/sigir/TayPLH17} then learns a similarity score between document pairs.

Recent advances in neural ranking models go beyond independent representation learning. There are several main architectural paradigms that invoke interactions between document pairs which intuitively improve performance due to matching at a deeper and finer granularity. The first can be thought of as extracting features from a constructed word-by-word similarity matrix \cite{DBLP:conf/aaai/WanLGXPC16,pang2016text}. The second invokes matching across multiple views and perspectives \cite{DBLP:conf/emnlp/HeGL15,DBLP:conf/ijcai/WangHF17,1805.08159}. The third method involves learning pairwise attention weights (i.e., co-attention). In these models, the similarity matrix is used to learn attention weights, learning to attend to each document based on its partner. Attentive Pooling Networks \cite{DBLP:journals/corr/SantosTXZ16} and Attentive Interactive Networks \cite{DBLP:conf/aaai/ZhangLSW17} are models that are grounded in this paradigm, utilizing \textit{extractive max-pooling} to learn the relative importance of a word based on its maximum importance to all words in the other document. The Compare-Aggregate model \cite{DBLP:journals/corr/WangJ16b} used a co-attention model for matching and then a convolutional feature extractor for aggregating features. Notably, other related problem domains such as machine comprehension \cite{DBLP:journals/corr/XiongZS16,seo2016bidirectional,wang2016machine} and review-based recommendation \cite{tay2018multi} also extensively make use of co-attention mechanisms.

Learning sequence alignments via attention have been also popularized by models in related problem domains such as natural language inference \cite{DBLP:conf/emnlp/ParikhT0U16,DBLP:conf/acl/ChenZLWJI17,tay2017compare}. Notably, MCAN can be viewed as an extension of the CAFE model proposed in \cite{tay2017compare} for natural language inference. However, the key differences of this work is that (1) the propagated features in MCAN are \textit{multi-casted} (e.g., multiple co-attention variants are used consecutively) and (2) MCAN is extensively evaluated on a different and diverse set of problem domains and tasks.

There are several other notable and novel classes of model architectures which have been proposed for learning to rank. Examples include knowledge-enhanced models \cite{wu2016knowledge,xiong2017word}, lexical decomposition \cite{wang2016sentence}, fused temporal gates \cite{1711.07656} and coupled LSTMs \cite{DBLP:conf/emnlp/LiuQZCH16}. Novel metric learning techniques such as hyperbolic spaces have also been proposed \cite{Tay:2018:HRL:3159652.3159664}. \cite{zhang2018end} proposed a quantum-like model for matching QA pairs.

Our work is also closely related to the problem domain of ranking for web search, in which a myriad of neural ranking models were also proposed \cite{shen2014latent,shen2014learning,mitra2017learning,dehghani2017neural,hui2017pacrr,huang2013learning,hui2017re,xiong2017neural}. Ranking models for multi-turn response selection on Ubuntu corpus was also proposed in \cite{wu2016sequential}.

\section{Conclusion}
We proposed a new state-of-the-art neural model for a myriad of retrieval and matching tasks in the domain of question answering and conversation modeling.
Our proposed model is based on a re-imagination of the standard and widely applied neural attention. For the first time, we utilize attention not as a pooling
operator but as a form of feature augmentation. We propose three methods to compress attentional matrices into scalar features. Via visualisation and qualitative analysis, we show that these casted features can be interpreted and understood. Our proposed model achieves highly competitive results on four benchmark tasks and datasets. The achievements of our proposed model are as follows: (1) our model obtains the highest performing result on the well-studied TrecQA dataset, (2) our model achieves $9\%$ improvement on Ubuntu dialogue corpus relative to the best exisiting model, and (3) our model achieves strong results on Community Question Answering and Tweet Reply Prediction.
\bibliographystyle{ACM-Reference-Format}
\balance
\bibliography{./tybib/references}

\end{document}